\documentclass[english]{ccdconf}
\usepackage{amsmath}
\usepackage[scaled=1.0]{helvet}
\usepackage{times}
\usepackage{graphicx}
\usepackage{subfigure}
\usepackage{parskip}
\usepackage{multirow}
\begin{document}

\title{Single image dehazing via combining the prior knowledge and CNNs}
\author{Yuwen Li\aref{amss,amss1}, Chaobing Zheng\aref{hit}{*},
      Shiqian Wu\aref{hit}, Wangming Xu\aref{hit}}

\affiliation[amss]{Hubei Key Laboratory of Mechanical Transmission and Manufacturing Engineering, Wuhan University of Science and Technology, Wuhan 430081, China
        \email{liyuwen1205@163.com}}
\affiliation[hit]{The Institute of Robotics and Intelligent Systems, School of Information Science and Engineering, Wuhan University of Science and Technology, Wuhan 430081, China
        \email{zhengchaobing@wust.edu.cn, shiqian.wu@wust.edu.cn}, xuwangming@wust.edu.cn}
\affiliation[amss1]{School of Mechanical \& Electrical Engineering, Nanchang Institute of Technology, Nanchang 330099,China
        \email{liyuwen1205@163.com}}

\maketitle

\begin{abstract}
Aiming at the existing single image haze removal algorithms, which are based on prior knowledge and assumptions, subject to many limitations in practical applications, and could suffer from noise and halo amplification. An end-to-end system is proposed in this paper to reduce defects by combining the prior knowledge and deep learning method. The haze image is decomposed into the base layer and detail layers through a weighted guided image filter (WGIF) firstly, and the airlight is estimated from the base layer. Then, the base layer image is passed to the efficient deep convolutional network for estimating the transmission map. To restore object close to the camera completely without amplifying noise in sky or heavily hazy scene, an adaptive strategy is proposed based on the value of the transmission map. If the transmission map of a pixel is small, the base layer of the haze image is used to recover a haze-free image via atmospheric scattering model, finally. Otherwise, the haze image is used. Experiments show that the proposed method achieves superior performance over existing methods.
\end{abstract}


\footnotetext{* contributed equally. This work is supported by Wuhan University of Science and Technology Graduate Innovation and Entrepreneurship Fund Project(Grant No.JCX201973), National Natural Science Foundation of China (Grant No.61775172), Hubei Provincial Department of Education Research Program Project (Grant No.D20191104), The Science Foundation of Education Department of Jiangxi Provincial (Grant No.GJJ201927)} 

\section{INTRODUCTION}

Hazing images are usuall suffering from low contrast, color distortion and blurring, which will bring a lot of inconvenience to many outdoor computer vision applications such as video surveillance, smart transportation, aerial photography, and so on \cite{1ligq2013}. In recent years, single image haze removal has made a significant breakthrough, due to the reasonable prior knowledge and assumptions. At present, the image dehazing algorithms can be mainly divided into three types. Non-model-based image dehazing methods, model-based image dehazing methods and deep learning-based image dehazing methods. 

The non-model based image dehazing algorithm is based on human visual perception, enhancing the image contrast and correcting the color contrast to improve the quality, directly. Various Non-model methods have been applied to the problem of removing haze from a single image, including histogram-based \cite{Xu}, contrast-based\cite{Tan}, Homomorphic filtering\cite{Yu}, Retinex and improved Retinex based\cite{Retinex,Fan}. For example, Fan et al.\cite{Fan} proposed an image dehazing algorithm based on Retinex. These methods can make the image colors more balanced and softer, but the obtained image cannot be effectively enhanced in contrast, weakening the dark or bright areas in the original image, and blurring the focus of the image.  

The model-based image dehazing algorithms are based on the atmospheric scattering model, which can be used to recover the vivid image as much as possible. In \cite{ACM}, Independent Component Analysis (ICA) based on minimal input is proposed to remove the haze from color images, but the approach is time-consuming and cannot be used to deal with dense-haze images. He et al. \cite{He} estimated atmospheric light by dark channel prior and recovered the haze free image by atmospheric scattering model. Although the above algorithms have made great progress, they rely on a variety of prior knowledge, so they have limitations and not robustness. Inspired by an observation in \cite{He} that single image haze removal can be regarded as a type of spatially varying detail enhancement, a neat framework was proposed in \cite{Li2015} by introducing a local edge-preserving smoothing based method to estimate the transmission map of a haze image. These methods tend to over-estimate the haze concentration, resulting in excessive dehazing, especially in the sky region, which tends to cause color distortion, and the noise is also amplificated.

The third type of hazing removal method is based on deep learning. Deep learning is widely applied to address many image processing problems including low-light image enhancement \cite{LLnet,LightCNN,LDSI,Zheng2021}, single image rain removal \cite{Derain1,Derain2}, single image denoising \cite{denoise1,denoise2}, single image super-resolution \cite{VDSR, SRCNN}. It has also been widely used in the single image haze removal and achieved desired results to some extent. Ren et al. \cite{Multi} proposed a multi-scale deep neural network to estimate the transmission map. But the network was training by minimizing the loss between the reconstructed transmission mapping and the corresponding ground truth mapping, not the haze removal image and the ground truth image. It can achieve desired results in daytime hazy images but not nighttime hazy images. Cai et al. \cite{cai} also proposed using a convolutional neural network to learn the transmission map. However, due to the particularity of some scenes, the estimation of the transmission map is not accurate, the effect of dehazing is not ideal. When the depth of field transition in the image is large, the effect of dehazing is not ideal for distant objects, especially in the sky area. This is mainly due to the shallow network, which cannot well learn the mapping relationship between the haze image and the corresponding ground truth image, and some detail loss. Qu et al. \cite{Qu_2019_CVPR} proposed an Enhanced Pix2pix Dehazing Network (EPDN), which generated a haze-free image without relying on the physical scattering model. It can obtain a haze-free image with faithful color and rich details. But it is not very robust for heavily hazy scene, the edges of objects in heavily haze cannot be recovered naturally, especially in the sky region.

Although the above methods have made great progress in a single image dehazing, they still have some defects. It could suffer from noise amplification in the sky region and possible color distortion in the restored image. In this paper, a new single image dehazing via combining the prior knowledge and CNNs is introduced to deal with it. The haze image is decomposed into a base layer and a detail layer by using an edge-preserving smoothing filter such as the weighted GIF (WGIF) \cite{Li2014}, firstly. As we know, noise is mainly included in the detail layer. Then, the airlight and transmission map are both estimated from the base layer to reduce the effect of noise. The airlight is estimated from the base layer by using a hierarchical searching method based on the quad-tree sub division \cite{A}. The base layer image is passed to the efficient deep convolutional network for estimating the transmission map. The atmospheric scattering model is used to recover the haze-free image, finely. The network is training by minimizing the loss function between the haze removal image and the ground truth image, so it is more robust. To reduce the noise in sky or heavily hazy scene, and restore the objects close to the camera completely, another strategy is cleverly used. If the transmission map of a pixel is small, it usually belongs to sky or heavily hazy scene, only the base layer of the haze image is used to recover a haze-free image via atmospheric scattering model, the noise is not amplified in the final image. Otherwise, the haze image is used. As such, the possible effect of noise is reduced significantly in the restored image. In summary, our contributions are highlighted as follows:

\begin{itemize}
	\item The influence of noise in the existing methods is analyzed in this paper detailly. A clever strategy is used to reduce noise interference in sky or heavily hazy scene.
	\item Single image dehazing via combining the prior knowledge and CNNs is proposed in this paper. The haze image is decomposed into a base layer and a detail layer, and only the base layer is used to estimated airlight and transmission map to reduce the effect of noise.
	\item The network is training by minimizing the loss function between the ground truth image and haze removal image recovered via atmospheric scattering model. The loss function contains both structural information and color information in this paper.
\end{itemize}

The remainder of this paper is organized as follows: the preliminary knowledge on single image haze removal are presented in sections II, our method for single image dehazing is proposed in Section III. Experimental results are provided in Section IV to verify the proposed method. The conclusions are drawn in Section V, finally.

\section{EFFECT OF NOISE ON IMAGE HAZE REMOVAL}
Image dehazing means recovering clear images from a noisy frame caused by haze, fog or smoke, and the atmospheric scattering model is often used to deal with it. It can be formally written as:
\begin{equation}
\label{eq1}
Z(p) = I(p)t(p) + A(1-t(p))
\end{equation}
Where $p$ is the pixel position, $Z$ is a haze image, $I$ is a haze-free image, $t$ is the transmission map describing the portion of the light that is not scattered and reaches the camera, and $A$ is the global airlight. Many methods have been already proposed to estimated $t$ and $A$, such as the dark channel prior algorithm \cite{He}, deep learning algorithm \cite{Dehaze} and so on. When both  $t$ and $A$ are known, then the haze free can been recovered by:
\begin{equation}
\label{eq2}
I(p) = \frac{{Z(p)-A(p)}}{max(t(p),{t_0})}+A(p)
\end{equation}
Where $t_0$ is a constant to preserve a small amount of haze in dense haze regions \cite{He}. However, noise is very common in haze images, these methods cannot eliminate the effects of noise well. Although some previous work \cite{Li2014,Li2018,Li2015} have already been done to eliminate the effects of noise in sky or heavily hazy scene, they did not specify how noise affects image dehazing. Some analyses will be given, then an effective solution will be presented in this section. 

If the haze image $Z$ with noise can be presented as: 
\begin{equation}
\label{eq31}
Z(p) = {\tilde{Z}}(p) + n(p)
\end{equation}
Where $\tilde{Z}$ presents the haze-free image without noise, $n$ is the noise. From the Eq.  (\ref{eq1}) and (\ref{eq31}), the image dehazed can be expressed as:
\begin{equation}
\label{eq4}
I(p) = \frac{{Z(p)-A(p)}}{max(t(p),{t_0})}+A(p)+ \frac{{  n(p)  }}{max(t(p),{t_0})}
\end{equation}

\begin{figure}[htb]
	\centering
	\includegraphics[width=0.45\textwidth]{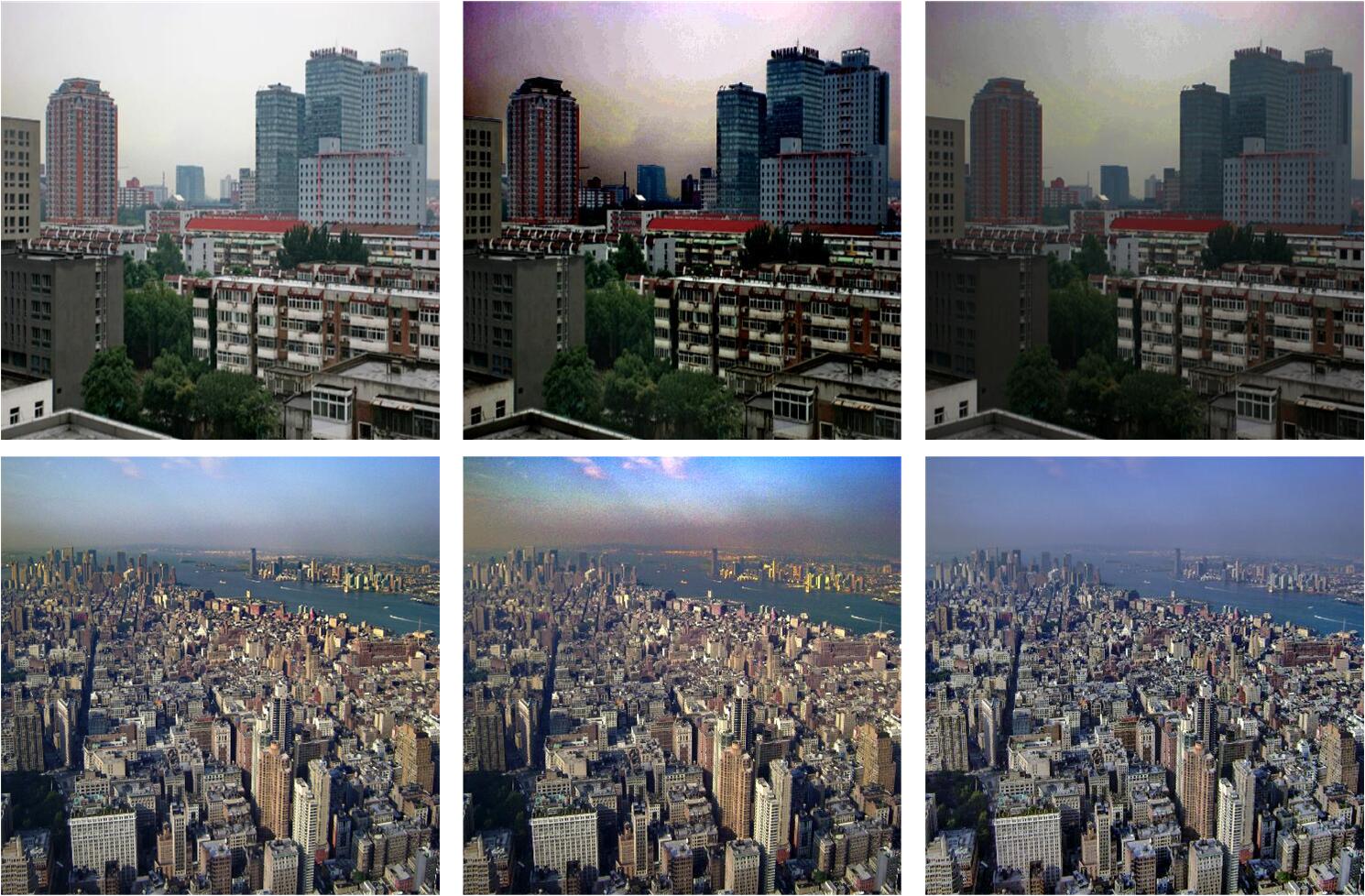}
	\caption{The first column are hazing-free images. The other two columns are the results of dehazing. The noise is amplificated in the sky regions.}
	\label{Fig1}
\end{figure}

When $t(p)$ is close to zero, $max(t(p),{t_0})$ is smaller than 1, the point $p$ belongs to sky or heavily hazy scene [He], the noise is amplificated as $\frac{{  n(p)  }}{max(t(p),{t_0})}$, as shown in Eq. (\ref{eq4}),some dehazing results are inshown in Fig. \ref{Fig1}.

\begin{figure*}[htb]
	\centering
	\includegraphics[width=0.95\textwidth]{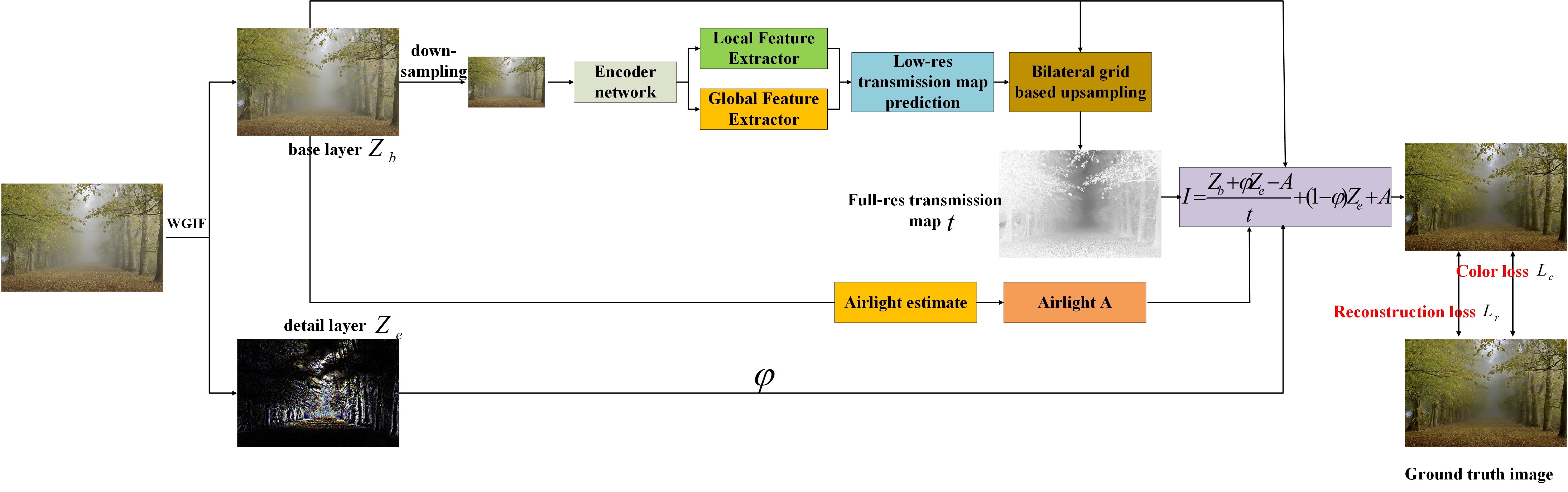}
	\caption{Overview of our method. First, The haze image is decomposed into base and detail layer by using WGIF. Then, the base layer is used to estimated $t$ and $A$ by using the network. $Z_b$ is pass to the network after downsampling to predict the low-res tranmission map. Finally, we upsample the result to produce the full-res tranmission map $t$, and take it to recover the haze removal image.}
	\label{Fig2}
\end{figure*}

It can see from the Fig. 1 that noise in sky or heavily haze scene are all amplificated, and color is distorted, which negatively influences the visual of the pictures. Therefore, an effective measure needs to be put to address this problem.

In addition, the global airlight is affected by noise. For example, $A$ is often estimated by using a hierarchical searching method based on the quad-tree subdivision \cite{A} from the haze image $Z(p)=\tilde{Z}(p)+n(p)$. And the $A_c$ can be presented  as $\tilde{Z}(p_0)+n(p_0)$, $p_0$ can be obtained by minimiizing the distance $\parallel \tilde{Z}_c(p)+n_c(p) -255  \parallel$. 

Furthermore, the estimation of $t$ is also affected by noise. Many single image degazing methods are based on dark channel prior \cite{He}, $t$ can be presented as:
\begin{equation}
\begin{aligned}
\label{eq5}
t(p) &= 1-w{\ast}\mathop{min}\limits_{c}(\mathop{min}  \limits_{p' \in \Omega(p) }  (\frac{{Z_c(p)}}{A_c}))\\          &=1-w{\ast}\mathop{min}\limits_{c}(\mathop{min}  \limits_{p' \in \Omega(p) } (\frac{{\tilde{Z}_c(p)+n_c(p)}}{A_c} )   )
\end{aligned}
\end{equation}
Clearly, the estimation of $t$ is inaccurate from the Eq. (\ref{eq5}). Another common ways to estimate $t$ are deep learning \cite{Dehaze,Map2}. Learn a mapping function $f(\cdot)$ through haze images directly, then $t$ can be presented as:
\begin{equation}
\label{eq6}
t(p) =f(Z(p)) =f(\tilde{Z}(p)+n(p))
\end{equation}
Unfortunately, the noise in $Z$ is random, and it is difficult to obtain its regular pattern, so the $f(\cdot)$ is not precise enough.

%

%
%
%
%
%
%

\textbf{\section{PROPOSED METHOD}}
As discussed above, the haze image could suffer from noise. It cannot correctly been described by the imaging model. Therefore, an end-to-end network framework is expected to be proposed, which takes a hazy image as input, and the outputs its haze-free image. However, this mapping function is difficult to learn and acquire, which is easy to cause color deviation, and icrease the burden of learning \cite{Fu2017RemovingRF}. An end-to-end haze removal algorithm based on prior knowledge and deep learning is proposed in this section.
Mathematical formulas should be roughly centered and have to be numbered
as formula~(\ref{eq1}).

\subsection{Framework of the single image dehazing}

Fig. \ref{Fig2} shows the framework of our algorithm, it has three major advantagers. First, it is robust to noise. The input of the network is $Z_b$, the outputs is $t$ that is subsequently used to recover a haze-free image via atmospheric scattering model. And $A$ is also estimated from the $Z_b$. Second, it is very effective. An end-to-end system combining the prior knowledge and deep learning is designed in this paper. Both color and reconstruction loss are taken into account for learning the transmission mapping from the haze image and ground truth image, it is more robust. Third, it is efficient. As shown in Fig. \ref{Fig2}, most nework computation is done in low-res domain, then bilateral grid based upsampling is adopted to obtain full-res transmission map. It is friendly to hardware with insufficient computing power.

Firstly, WGIF is adopted to decomposed the haze image into base and detail layer. The former is defined as:
\begin{equation}
\label{eq3}
Z(p) = Z_e(p) + Z_b(p)
\end{equation}
Where $Z_e$ and $Z_b$ are the detail and base layer of the haze image $Z$. As we known, noise is included in the detail layer, so both $t$ and $A$ are estimated from $Z_b$ rather than $Z$ same as in \cite{Li2014}. Same to \cite{A,Li2014}, a hierarchical searching method based on the quad-tree subdivision is used to estimated $A$. 
Let $a$ and $b$ are coefficients of the WGIF, They can been obtained by:
\begin{align}
\left\{\begin{array}{l}
a_{p} = \frac{\sigma^2_{Z, \zeta}(p)}{\sigma^2_{Z, \zeta}(p)+ \frac{\lambda}{{\Gamma_Y}(p)} }\\
b_{p} = (1-a_{p})\mu_{Z,\zeta}(p)
\end{array}
\right..
\end{align}
Where $\mu_{Z,\zeta}(p)$ and $\sigma^2_{Z, \zeta}(p)$ are the mean value and the covariance value of $Z$ in the windows of $\Omega_{\zeta}(p)$, the weight of ${\Gamma_Y}(p)$ as fellows:
\begin{eqnarray}
\label{w(p)}
{\Gamma_Y}(p)=\sum_{p'=1}^{N}\frac{\sigma^2_{Y,1}(p)+\varepsilon}{\sigma^2_{Y,1}(p')+\varepsilon},
\end{eqnarray}
and $Y$ is the luminance of $Z$. The base layer $Z_b$ can been obtained by:
\begin{equation}
{Z_b}(p)=\bar{a}{Z(p)}+\bar{b}_p,
\end{equation}
Where $\bar{a}$ and $\bar{b}$ are the means of $a_p$ and $b_p$ in the windows of $\Omega_{\zeta}(p)$. Inspired by \cite{Du,hdrnet}, HDRNet \cite{hdrnet} is adopted to obtain the $t$ not the haze-free image. It has been discussed in \cite{hdrnet} that HDRNet is not suitable for dehazing directly, it could reduce local contrast and destroy image detail. Because it breaks the assumption that the relationship between the input and output should be a local affine transformation. Assuming that $Z$ and $I$ are available when training the network, $A$ can be estimated by \cite{A}, then the relationship between $I$ and $t$ can be regarded as a local affine transformation from the Eq. (\ref{eq1}). So HDRNet can be used for estimating $t$.

When $A$ and $t$ have been estimated, Eq. (\ref{eq2}) can be used to obtain dehazing images. If the hazing image $Z$ is used directly, all the noise will be amplified \cite{He}. However, if only the base layer image $Z_b$ is used for dehazing, some details will be lost. In order to reduce the influence of noise while retaining the details, a signal weight $\psi$ is adopted to determine whether to decompose or not based on the value of $t$.  The former is defined as:
\begin{equation}
\label{eq8a}
\psi(t) =  \frac{1}{ 1 + e^{32(1-{\eta}{t} )}}
\end{equation}
Where $\eta$ is a constant. if $t$ is smaller than $1 {/} {\eta}$, the pixel belongs to very dense haze regions or sky area, only the base layer is uesd to recover the dehazing image. Otherwise the original pixel in $Z$ is uesd directly. Then, the final image can be obtained by:

\begin{equation}
\begin{aligned}
\label{eq9a}
I(p) = \frac{Z_b(p) + {\psi}(t(p)) {Z_e}(p) - A(p) }{t(p)} \\
   + (1-{\psi}(t(p)) ){Z_e}(p) + A(p)
\end{aligned}
\end{equation}

From the Eq. (\ref{eq9a}), when $t(p)$ is smaller than $\eta$, the pixel in position $p$ belongs to the very dense haze regions or sky area, and $\psi(t)$ is $0$, only the base layer is amplified. $t(p)$ is usually larger than $\psi(t)$ when the pixel belongs to the object nearby the camera, the detail of the pixel is enhanced. So we can obtain the dehazing image without amplifing the noise in sky or dense haze regions.

\subsection{Loss Function of Our Framework}

In order to learn the transmission map from a set of haze and haze-free images pairs, the loss function $L$ used in this paper takes into account the consistency of color and structure. It is presented as:
\begin{equation}
\label{eq8}
L =L_r + w_c \ast L_c
\end{equation}
Where $L_r$ is reconstruction loss, $L_c$ is color loss, $w_c$ is a constant, and its value is selected as 0.01 in this paper.

The restoration loss $L_r$ is usually defined as
\begin{equation}
L_r = \sum_{p,c}[\tilde{I}_{c}(p)-I_c(p)]^2.
\end{equation}
Where $\tilde{I}$ is haze-free image, ${I}$ is dehazing image which can be obtained by Eq. (\ref{eq9a}).
As we known, the results of CNNs look a litter blurry \cite{pix2pix}. In order to measure the color difference between the dehazing images and target images, we propose applying a Guassian blur and Cosine distance between the obtained representations. Blurring removes high-frequencies and makes color comparison easier. It is suitable for the measurement of color information of the reaults of CNNs. The color loss $L_c$ is defined as:
\begin{equation}
\label{a1}
L_c=\sum_{p,c}\angle ( \tilde{X}_{c}(p), Y_c(p)   ),
\end{equation}
Where $\tilde{X}$ and $Y$ are the blurred images of $\tilde{I}$ and $I$, respectively.
\begin{equation}
\label{a2}
\tilde{X}_{c}(p_i,p_j)=\sum_{k,l}X({p_i}+k,{p_j}+l) \cdot G(k,l),
\end{equation}
and the 2D Gaussian blur operator is given by
\begin{equation}
\label{a3}
G(k,l)= A  \cdot exp(-\frac{(k-{\mu}_x)^2}{2{\sigma}_x} - \frac{(l-{\mu}_y)^2}{2{\sigma}_y}  )
\end{equation}
Where we defined $A=0.053$, ${{\mu}_{x,y}}=0$, and ${{\sigma}_{x,y}}=3$.  $\angle (\tilde{I}_{c}(p), I_c(p))$ is the angle between two $3D(R,G,B)$ vectors $\tilde{I}_{c}(p)$ and $I_c(p)$. Although $L_r$ can measure the similarity of the two vectors numerically, it cannot make sure that their color vectors are in same direction.

\section{EXPERIMENTAL RESULTS}
Extensive experimental results are provided in this section to validate the proposed framework in dehazing.

\subsection{Dateset}
We evaluated our framework in NYU-Depth \cite{NYU}. NYU-Depth consists of 1499 pairs of hazing and clean images. 1399 pairs of images of NYU-Depth and pictures we collected are used for training the net, the remaining 100 pictures will be used for testing. Mirroring, cropping are employed to augment data.  The images are cropped to $480\times480$ and downsampled to $256\times256$ and sent to the network, then the bilateral grid-based module is used to upsample the $t$ to reduce the computational cost.

\subsection{Ablatiom Study on Loss Functions}
\begin{figure}[htb]
	\centering
	\includegraphics[width=0.45\textwidth]{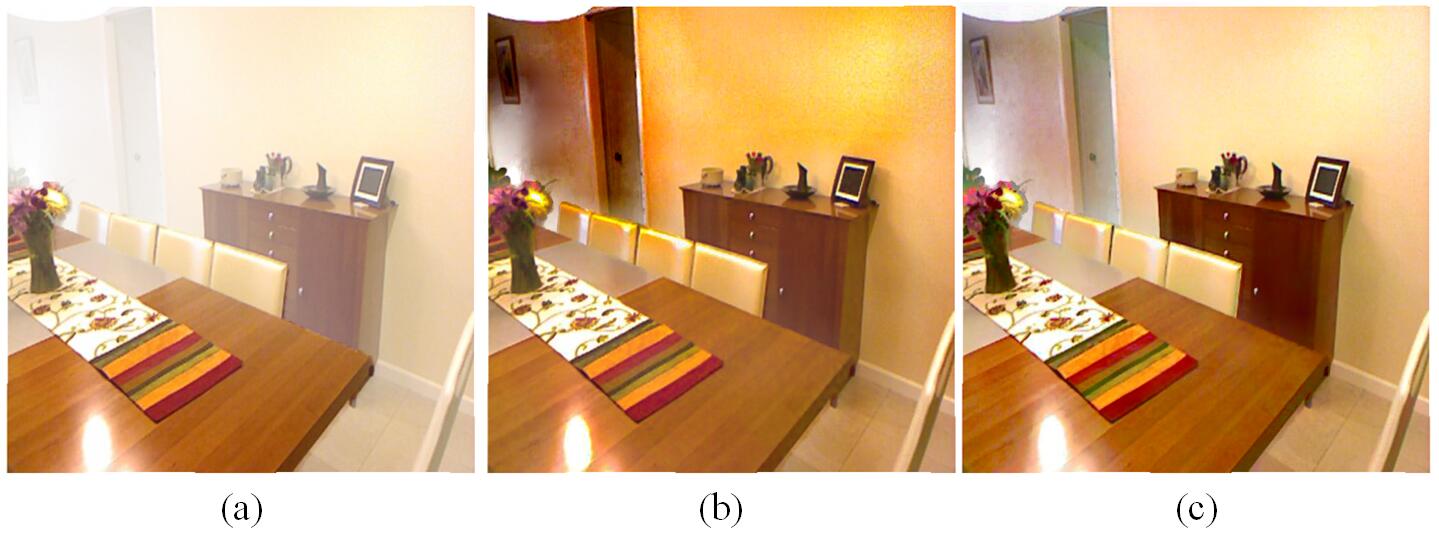}
	\caption{ Comparison of different lossfunction. From left to right, hazing image, restored images by using $L_r$ and $L_r + L_c$ }
	\label{Fig3a}
\end{figure}

As we konwn, if we take a native approach and ask the CNN to mininize the Euclidean distance between predicted and ground truth pixels, it will tend to produce blurry results \cite{pix2pix}. Beacuse Euclidean distance is minimized by averaging all plausible outputs, which causes blurring. So color loss function is added to ensure the dehazing images and ground truth images vectors have the same direction, some results are shown in Fig.\ref{Fig3a}. The result of $L_r$ is dark, and some detail information is loss.

\begin{figure*}[htb]
	\centering
	\includegraphics[width=0.95\textwidth]{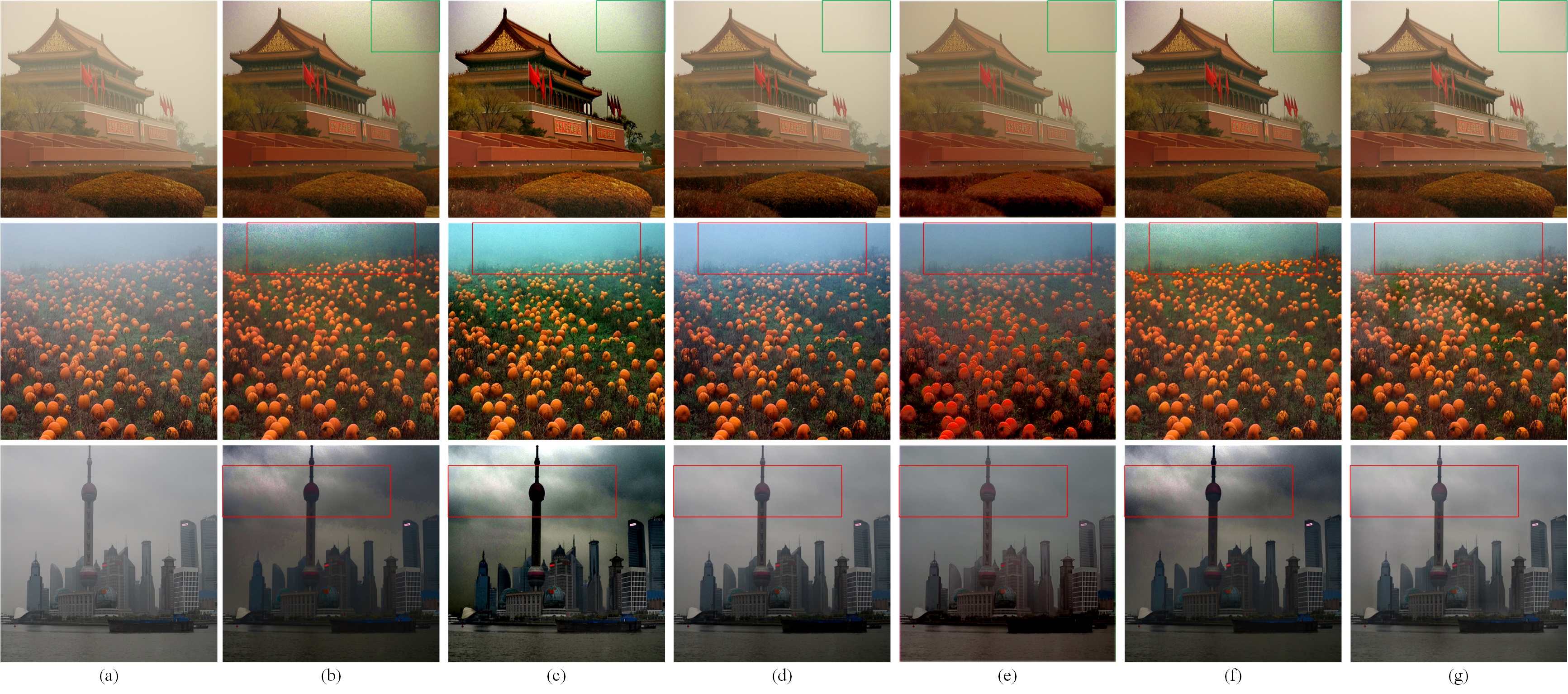}
	\caption{ Comparative results of the proposed algorithm and other five metgods on outdoor images. (a) The hazing images; (b) The dehazing images by \cite{He}; (c) The dehazing images by \cite{CVPR16}; (d) The dehazing images by \cite{Dehaze}; (e) The results of \cite{ICCV17};  (f) The results of \cite{YMM}; (g) The results of the proposed algorithm;  }
	\label{Figa}
\end{figure*}
\begin{figure*}[htb]
	\centering
	\includegraphics[width=0.95\textwidth]{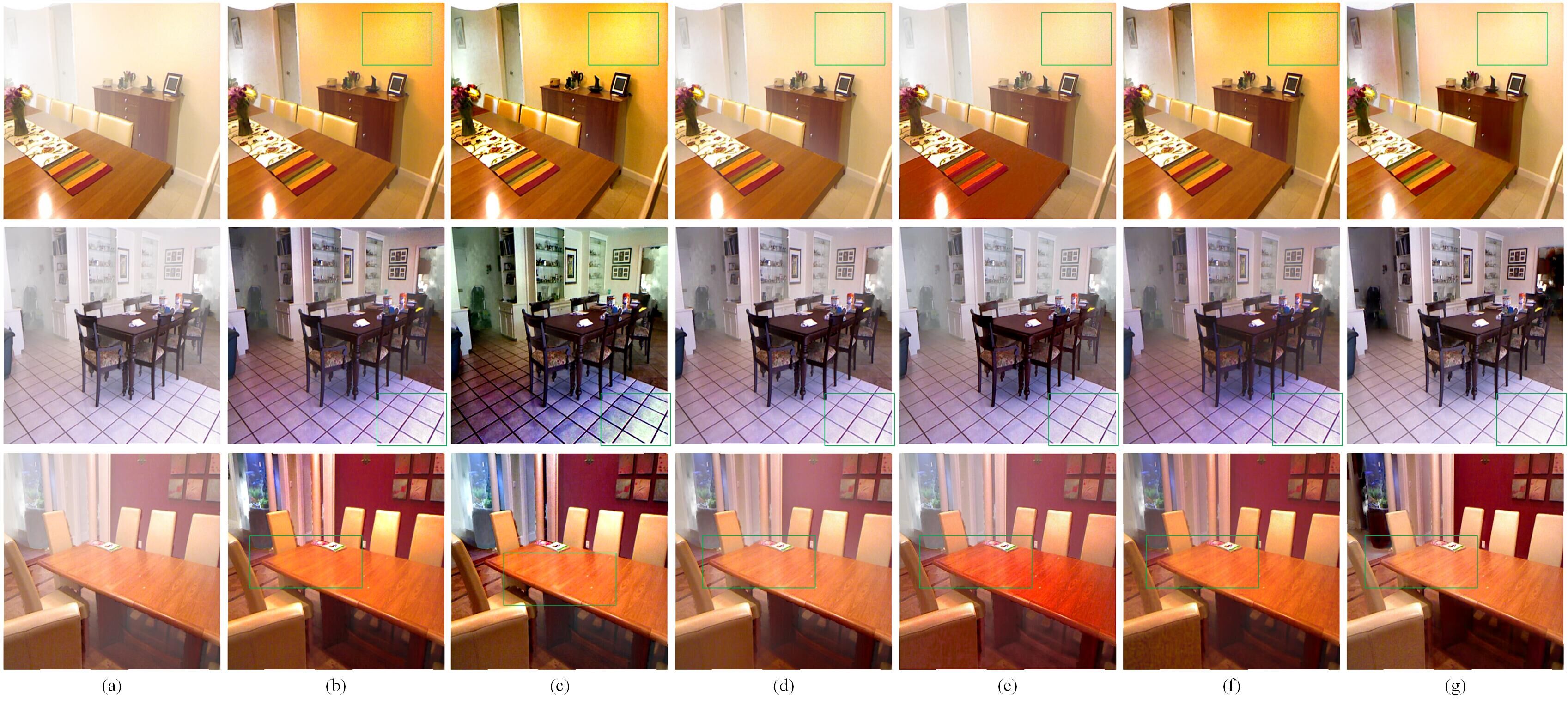}
	\caption{ Comparative results of the proposed algorithm and other five metgods in NYU-Depth. (a) The hazing images; (b) The dehazing images by \cite{He}; (c) The dehazing images by \cite{CVPR16}; (d) The dehazing images by \cite{Dehaze}; (e) The results of \cite{ICCV17};  (f) The results of \cite{YMM}; (g) The results of the proposed algorithm; Noise in results of \cite{He, CVPR16, Dehaze, ICCV17} is amplified, and the colors is distorted.}
	\label{Figb}
\end{figure*}

\subsection{Comparison with State-of-the-art Methods}
In order to demonstrate the efficiency of our method, we compare it with five state-of-art dehazing algorithms in this subsection. Some results are shown in Fig. \ref{Figa} and \ref{Figb}, the results of \cite{CVPR16,He} have severe color distortion, and noise is also amplificated, especially in the sky regions. The results by \cite{YMM} look a little dark and unreal, and the results of \cite{Dehaze, ICCV17} look a litter blur. The results of the proposed look much sharper and more 
realistic, details have also been enhanced and without color distortion. Readers are invited to view to electronic version of full-size figures and zoom in these figures so as to better appreciate differences among images

The subjective evaluation has been performed above, and then SSIM and PSNR will be adopted for objective comparison. As shown in Table \ref{tab1}, the proposed method generates the results with higher PSNR and SSIM values than those of other algorithms.

{\small \renewcommand{\arraystretch}{1.2}
\begin{table}[!htb]
	\setlength{\abovecaptionskip}{0.cm}
    \setlength{\belowcaptionskip}{-0.cm}
	\caption{The results of SSIM on the NYU-Depth}
	\label{tab1}
	\begin{tabular}{c|cccccp{12pt}c}
		\hhline
		     &\multirow{2}*{\cite{He}}& \multirow{2}*{\cite{CVPR16}}& \multirow{2}*{\cite{Dehaze}}& \multirow{2}*{\cite{ICCV17}}& \multirow{2}*{\cite{YMM}}  & Ours (no color)   & \multirow{2}*{Ours} \\ \hline
		SSIM & 0.798 & 0.701 &0.772 &0.717 &0.796 &0.795 &\textbf{0.800} \\ 
		\hline
		PSNR & 16.24  & 14.28  &14.62  &13.24  &15.95  &17.69 &\textbf{17.74}   \\
		\hhline
	\end{tabular}
\end{table}}

\begin{table}[htb]
	\begin{center}
		\centering
		\caption{The results of SSIM on the NYU-Depth}
		\tabcolsep8pt\begin{tabular}{cccc}
			\hline		
			\multirow{1}*{ }   &   SSIM    &  PSNR\\
			\hline
			\multirow{1}*{\cite{He}} & 0.9807 & 40.9830\\
			\multirow{1}*{\cite{CVPR16} } & 0.9768 & 40.0619\\
			\multirow{1}*{\cite{Dehaze}} & 0.9823 & 41.5068\\
			\multirow{1}*{\cite{ICCV17} } & 0.9831 & 41.5606\\
			\multirow{1}*{\cite{YMM} } & 0.9828 & 41.5521\\
			\multirow{1}*{Ours (no color) } & 0.9838 & 17.69\\
			\multirow{1}*{Ours } & \textbf{0.9838} & \textbf{17.74}\\
			\hline
		\end{tabular}
		\label{tab1}
	\end{center}
\end{table}

\section{CONCLUSION}
A single image dehazing via combining the prior knowledge and deep learning is proposed in this paper. Firstly, the hazing image is decomposed into base layer and detail layer by weighted guided image filter. Then, the value of transmission map is estimated on the base layer by using deep learning method. To avoid the effects of noise more effectively, a weight function is designed to determine whether the image is decomposed or not. Finally, the atmospheric scattering model is used to restore the image.  The priori knowledge model and deep learning are cleverly combined, and adopt the advantages of both fully. Experimental results show that the proposed algorithm outperforms existing algorithms.

It should be pointed out that the proposed algorithm can not avoid the impact of noise completely as shown in Fig \ref{Figa}. The next R${\&}$D problem is to further reduce noise, especially dense hazing image or hazing image at night. Besides the dehazing, the secure communication \cite{1lik2003} is also important for real outdoor vision systems. All these problems will be studied in our future research.

\end{document}